\documentclass[conference]{IEEEtran}

\usepackage{cite}
\usepackage{amsmath,amssymb,amsfonts}
\usepackage{cite}
\usepackage{algorithmic}
\usepackage{graphicx}
\usepackage{textcomp}
\usepackage{xcolor}
\def\BibTeX{{\rm B\kern-.05em{\sc i\kern-.025em b}\kern-.08em
    T\kern-.1667em\lower.7ex\hbox{E}\kern-.125emX}}

\begin{document}

\title{Testing the effectiveness of saliency-based explainability in NLP using randomized survey-based experiments\\
}

\author{\IEEEauthorblockN{1\textsuperscript{st} Adel Rahimi}
\IEEEauthorblockA{\textit{Procter \& Gamble}\\
Singapore \\
rahimi.a@pg.com} 
\and
\IEEEauthorblockN{2\textsuperscript{nd} Shaurya Jain}
\IEEEauthorblockA{\textit{Singapore Management University} \\
Singapore \\
shauryaj.2019@economics.smu.edu.sg}

}

\maketitle
% \blfootnote{This publication is not on behalf of the Procter \& Gamble company.}

\begin{abstract}

\end{abstract}
As the applications of Natural Language Processing (NLP) in sensitive areas like Political Profiling, Review of Essays in Education, etc. proliferate, there is a great need for increasing transparency in NLP models to build trust with stakeholders and identify biases. A lot of work in Explainable AI has aimed to devise explanation methods that give humans insights into the workings and predictions of NLP models. While these methods distill predictions from complex models like Neural Networks into consumable explanations, how humans understand these explanations is still widely unexplored. Innate human tendencies and biases can handicap the understanding of these explanations in humans, and can also lead to them misjudging models and predictions as a result. We designed a randomized survey-based experiment to understand the effectiveness of saliency-based Post-hoc explainability methods in Natural Language Processing. The result of the experiment showed that humans have a tendency to accept explanations with a less critical view. 
\\  
\begin{IEEEkeywords}
AI Explinability, Machine Learning, Human-centric AI, XAI
\end{IEEEkeywords}
\section{Introduction}

Machine Learning projects in the past relied on human knowledge and were easily comprehensible by both developers and end-users. For example, decision trees have been widely used for fraud detection \cite{sahin2013cost}. However, the surge in popularity of Deep Neural Network (DNN) based models in the previous decade, due to their potential to produce more accurate results, has abated this comprehensibility. An example is the Convolutional Neural Network developed by Krizhevsky that was able to achieve ``top-1 and top-5 error rates of 39.7\% and 18.9\%"  \cite{krizhevsky2012imagenet} on Imagenet \cite{deng2009imagenet} comparing that to the second place error rate that was 26.2\% on the top-5 error rate. Such superior performance of DNNs has come at the cost of explainability as they can comprise up to millions or even billions of parameters \cite{brown2020language} \cite{shoeybi2019megatron} whose synergy in arriving at the final output is often undecipherable. Moreover, these models are considered to be ‘black boxes’, where not only the end-users but also the developers are unaware of their inner workings.

This reduced explainability has compromised the trust various stakeholders have in these models, despite their accurate results \cite{guidotti2018survey}. Transparency breeds trust in humans, and it is important to create transparency through explainability of these complex models \cite{pedreshi2008discrimination}. It is only becoming more vital as AI gains foothold in making critical – and in some cases, fatal – decisions in sensitive areas like Healthcare, Finance, Automated Driving, and such-like \cite{pasquale2015black} \cite{holzinger2019causability} \cite{babic2021beware}. The true potential of these recent advancements in AI can only be realised if the various stakeholders manage to discern the working of AI models and how their predictions are produced, as that is necessary to incorporate trust. For example, 83\% of people do not understand automated decision-making systems in the criminal justice system, and subsequently, 60\% oppose its use in the domain \cite{balaram2018public}. But besides securing the buy-in of end-users and developers through building trust, AI explainability also has the potential of identifying AI inaccuracies prior to deployment. This is crucial considering that black-box machine learning models have led to unjustified social ills, like unfavourable parole denials and credit decisions based on race \cite{larson2016recidevism}. 

This has led to the creation of Explainable AI (XAI), a subfield in AI which focuses on explaining AI decisions in human-understandable ways. Explainability itself is not a new concept \cite{xu2019explainable}; Explanations of AI systems have been used to debug AI systems already \cite{scott1977explanation}.

While model interpretability can be augmented through restricting their complexity, this can limit the effectiveness of the models \cite{dziugaite2020tradeoff}. As such, there has been an explosion of work around post-hoc analysis in XAI, whereby methods are devised to explain the models post-training without aiming to make the models inherently interpretable \cite{kenny2021post}. These recent developments are able to distil predictions from complex models like Neural Networks into simple explanations that can be consumed by both developers and end-users alike. However, this points the torch to the similarity between these explanations and human understanding \cite{hagras2018toward}. That is, whether these explanations of AI predictions and models are understood by humans as intended is still an open question. Innate human tendencies, biases, etc. can handicap the effectiveness of these explanations in humans accurately assessing predictions and models, and can even have the reverse of the intended outcome. This gap between explanation and understanding can be dangerous, especially in fields like healthcare, as it can lead to stakeholders considering incorrect models to be correct, and vice-versa. These necessitate more work to be done in this field.

The need for explainable AI is particularly important in Neural Network based NLP models — a field which is gaining foothold in making pertinent decisions, and hence, requires transparency. For example, NLP are in inchoate stages of being used for political-profiling for elections surveys \cite{malla2018political}. Hence, NLP can heavily influence actual election outcomes through assessing these surveys \cite{blaispolls}. NLP-based models are also being used at a large scale in education for essay scoring. An example of this is evaluation of essays \cite{autoscoring2022}. Such important applications make NLP a pertinent AI area for explainability.

\section{Background}
% What is explanation and what is an interpretation
There are many definitions delineating what ``interpretation" and ``explanation" mean in the context of AI. Motavan et. al have drawn a distinction between the two \cite{montavon2018methods}. They describe ``Interpretability" as the ability to ``mapping of an abstract concept into a domain that humans can make sense of". For instance, text, symbols, images, etc. can help humans interpret the content and concept that they represent.``Explainability", on the other hand, has been put forth as ``the collection of features of the interpretable domain that contribute to a certain decision" \cite{montavon2018methods}. An example of this would be a heatmap highlighting the pixels that contributed the most to the categorization of an image. With the increasing complexity of AI models, the focus of AI ethics has shifted from making models inherently interpretable to explaining their decisions.

% What are different explanation types 
% a taxonomy
Inherently Explainable AI models are white-box models which are transparent by design. By delineating the most important parameters considered, they provide an intuitive understanding of why the model has predicted a certain data point as belonging to a certain class. Some common examples of Inherently Explainable AI models are Decision Trees \cite{quinlan1986induction} and Linear Predictors. However, more complex problems might not be solved by Inherently Explainable AI models and require black-box models e.g. Deep Learning \cite{lecun2015deep} \cite{hochreiter1997long}.

The most popular methods for explainability can be summarized into three main categories spanning the entire cycle of modeling: Pre-modeling explainability \cite{sachan2021evidential}, Modeling explainability or explainable modeling \cite{gunning2017explainable}, and Post-modeling explainability \cite{samek2017explainable}. 
% add more citations here

Pre-modeling explainability focuses on the study of the input. It is the most rudimentary explainability method, and it comprises inspection of the input through plotting the data distribution, analyzing different classes, and even showing word tokens and class labels in the case of NLP. Such data exploration does not directly explain a model but provides useful insight into the model's behavior. For example, imbalance in the dataset can justify why the classification performance is subpar on underrepresented labels.

Modeling explainability methods rely on the model — rather than the data — to provide understanding to the end-user. These are the aforementioned Inherently Explainable AI models that are easy to understand due to the explicitness and the quantifiable number of parameters. Black-box models can not be explained through modeling-explainability methods.
% post-hoc XAI

A post-modeling or post-hoc XAI method produces approximations of black-box machine learning models in the form of simpler surrogate models, with a trained AI model as an input \cite{sara2018distill}. The approximations are constructed on the local behavior of a black-box model for a given input space. The main properties of such proxy models are their inherent explainability and their local faithfulness to the original model to be able to extrapolate its behavior. These simpler surrogate models can shed light on the decision logic of the black-box models through simple and understandable representations e.g. natural language, heatmaps, feature importance scores, etc. This can in turn assist human assessors in understanding and critiquing the black-box machine learning models after they are productionized. For example, the users can understand the importance of various features, highlight errors that the model is prone to, and discern biases in the model or data \cite{chandler2020bias}. Some well-known examples of these methods and packages are SHapley Additive exPlanations (SHAP) \cite{lundberg2020local} \cite{NIPS2017_7062}, LIME \cite{ribeiro2016should}, DeepLIFT \cite{shrikumar2017learning}, Interpret ML \cite{nori2019interpretml}, and AllenNLP Interpret \cite{wallaceallennlp}.

A number post-hoc methods derive feature importance by manipulating real samples, observing the change in the output given the changed samples, and generating a simple model that mimics the original model’s behavior in the original samples’ neighborhood. An example of such a method is Local Interpretable Model-agnostic Explanations (LIME) \cite{slack2020adverserial}. However, the aforementioned manipulation is done at random to produce the neighboring instances, and the local distribution of feature values and density of class labels in the neighborhood is not considered. The randomly generated instances used for the approximations may not even be observed in real samples. A class of methods relies on extracting decision sets that depict the decision logic of the Machine Learning model in the form of conditional rules\cite{lakkaraju2017approx}. Extracting a subgroup of rules from the copious amounts of decision rule possibilities is deemed as an optimization problem with classification accuracy and overall interpretability as two main objectives \cite{lakkaraju2016sets}. As building inherently interpretable models for complex and high-accuracy tasks is a challenge, post-hoc methods are amongst the most important means of explainability of Machine Learning models.

Despite this wide variety of post-hoc methods for different data types and different output types, there is no evidence to show that they realize their intended outcome of helping the user understand the internals and decision logic of complex models. While these models are able to distill complex black-box models into simpler models and undemanding representations, whether these simplifications translate into users accurately understanding the internals and logic of the model is uncertain. For example, the human tendency to fill in the blanks can lead them to manufacture additional information which misconstrues the explanation \cite{freeman1992cognitive}, or the anchor effect can hinder understanding when a misleading frame of reference is taken \cite{furham2011anchoring}. Thus, it is crucial to test the congruence between AI explanations and human understanding. In this paper, we test this on a very popular post-hoc method: Gradient-based saliency maps.

\section{Methodology}
% Explain how we design the form
A pre-trained model was used for the experiment: RoBERTa \cite{liu2019roberta} that was trained on Stanford Sentiment Treebank \cite{socher2013recursive}. The AllenNLP's Interpret library was used to generate the model's predictions and explanations \cite{wallaceallennlp} \cite{gardner2018allennlp}.

\begin{figure}[h]%
    \centering
    \includegraphics[width=\linewidth]{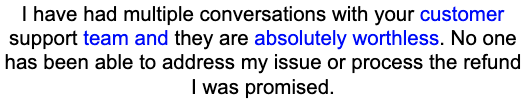}
    \caption{A negative classified review with top 5 discriminative words}
    \label{fig:saliency}
\end{figure}

Subsequently, ten product reviews from e-commerce website Amazon\footnote{https://www.amazon.com/} and restaurant reviews from travel review website Tripadvisor\footnote{https://www.tripadvisor.in/} were procured randomly, and then classified according to sentiment (positive or negative) using RoBERTa by AllenNLP. Based on the model card reported by AllenNLP\footnote{https://demo.allennlp.org/sentiment-analysis/roberta-sentiment-analysis}, the model has achieved 95.11\% accuracy in the test set, and hence, the predictions considered to be accurate. Subsequently, saliency map interpretations were generated for half of the predictions by visualising the gradient \cite{simonyan2013deep}. The text-specific class saliency maps highlighted the words in a given text that were discriminative with respect to the given class. These are post-hoc explanations to help users understand the top 5 words that helped RoBERTa in classifying the sentence into either negative or positive. An example of a review with top 5 discriminative words highlighted can be found in Figure \ref{fig:saliency}. In order to reduce the complexity of the research and the the top features were not, and were only highlighted without any specific order (as can be seen in \ref{fig:saliency}). This approach is very similar to the methodology in \cite{schuff2022human}, although we did not get the idea from the mentioned paper directly, we are excited that both approaches are similar. 

For the other half of the reviews, the words were highlighted at random instead of being based on the gradient. Keeping in mind that the randomly chosen word should not be one of the top-5 actual explanations. The simple gradient explanations and the ‘fabricated explanations’ – where the words were randomly highlighted –  were shuffled and put into the questionnaire alongside the prediction from the RoBERTa binary classifier. 

\begin{figure}[h]%
    \centering
    \includegraphics[width=\linewidth]{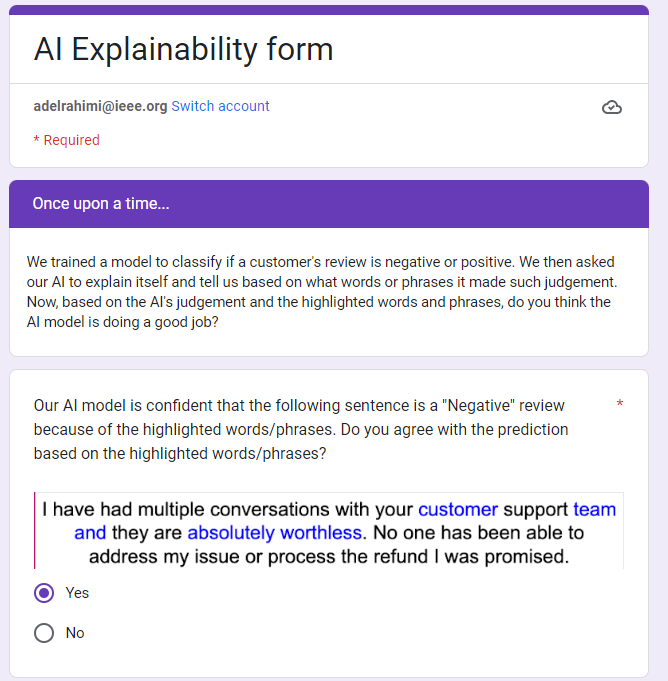}
    \caption{Screenshot of first question}
    \label{fig:question_screenshot}
\end{figure}

\begin{figure}[h]%
    \centering
    \includegraphics[width=\linewidth]{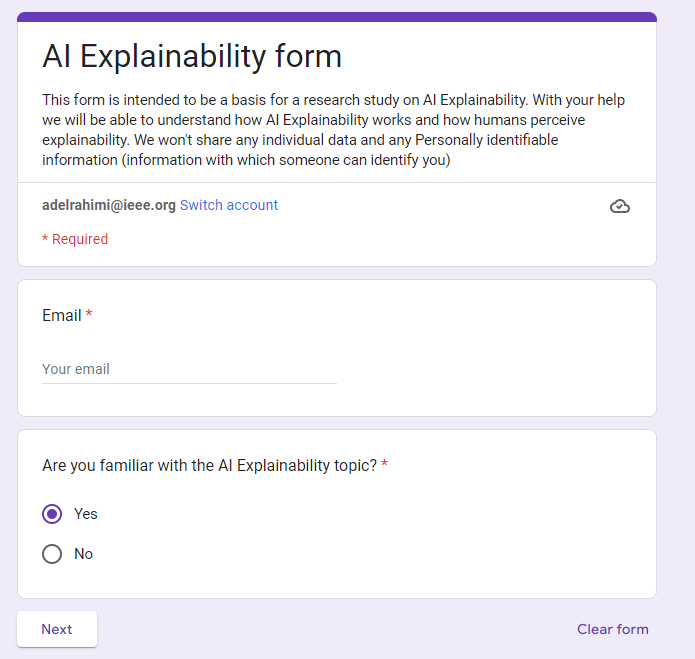}
    \caption{Screenshot of the survey}
    \label{fig:survey_screenshot}
\end{figure}

A survey was then created and distributed among 56 participants through email distribution lists. Figure \ref{fig:survey_screenshot} shows a screenshot of the survey's first page. For each of the reviews, the survey-takers were requested to assess whether the highlighted words in the explanation strongly suggest alignment with the classification (positive or negative). You can see a example in \ref{fig:question_screenshot}. If post-hoc explanations are effective in this scenario, the survey takers should ideally agree with the predictions of all the sentences with the original saliency map. Additionally, survey-takers would be expected to disagree with the predictions of all the sentences with the words highlighted at random.

As the questionnaire comprised 10 questions, a point was assigned to each question. If the participant was able to find out if the explanations were modified or not.

% to be edited

\section{Results}
% What were the results
 On average, the participants achieved a score of 6.43 out of 10. Meaning that on average, only 6.43 out of 10 times the participants agreed with the prediction with the unmodified explanation and disagreed with the prediction with the modified explanation. Table \ref{survey_stats_overall} shows the statistics from the participation survey.

\begin{table}[htbp]
\caption{Survey Statistics — Overall}
\begin{center}
\begin{tabular}{|c|c|c|c|}
\cline{1-4} 
\textbf{\textit{Average}} & \textbf{\textit{Median}}& \textbf{\textit{Range}}& \textbf{\textit{Mode}} \\
\hline
6.43 / 10                     & 6 / 10                     & 4 - 9                      & 5 / 10   \\
\hline
\end{tabular}
\label{survey_stats_overall}
\end{center}
\end{table}

Table \ref{survey_stats_unmod} shows the statistics of the responses to the predictions with unmodified simple gradient explanations. The points represent every time the survey taker agreed with an unmodified explanation's prediction. The high average score shows that survey-takers were mainly aligned with the predictions and the top-weighted words for classifying the sentences with unmodified explanations. The agreement with the predictions is expected due to the aforementioned high accuracy of the model used. However, the conformity of the survey-takers with the highest weighted words is indicative of the explanations accurately depicting the words that would lead to the sentence's classification. In other words, the simple gradient explanations in NLP are effective in helping end-users realise the most important features in predictions in a consumable form.

\begin{table}[htbp]
\caption{Survey Statistics — Unmodified explanations}
\begin{center}
\begin{tabular}{|c|c|c|c|}
\cline{1-4} 
\textbf{\textit{Average}} & \textbf{\textit{Median}}& \textbf{\textit{Range}}& \textbf{\textit{Mode}} \\
\hline
4.02 / 5                     & 4 / 5                      & 2 - 5                      & 4 / 5   \\
\hline
\end{tabular}
\label{survey_stats_unmod}
\end{center}
\end{table}

Table \ref{survey_stats_mod} shows the statistics of the responses to the predictions with modified simple  explanations. The points represent every time the survey taker disagreed with a modified explanation's prediction. The low average score indicates that the survey-takers often agreed with the randomised explanations justifying the respective prediction. This is counter to expectations, as the real highest-weighted discriminated words would often be missed, and words with low to negligible weights would often be highlighted as otherwise in the explanation. As such, these random words should not justify the prediction in most cases; however, the results show otherwise. We suspect that the participants draw improbable links between the explanations and the predictions. Seeing the false 'top-weighted words' and the prediction, they perhaps come up with ways that they can justify the prediction, which influences their perception of the explanation. This can create a gap between the intended effect of explanations and the actual understanding of explanations by the end-users and developers. If true, the effect of this tendency can make end-users assess incorrect/unfair models as correct/fair. Furthermore, this casts doubt on the effectiveness of unmodified explanations. In that, high number of agreements with original explanation predictions by the participants might have been driven by this tendency and not the actual effectiveness of the explanation. However, more work is required to further explore this hypothesis and other ones that could underpin this discrepancy. 

\begin{table}[htbp]
\caption{Survey Statistics — Modified explanations}
\begin{center}
\begin{tabular}{|c|c|c|c|}
\cline{1-4} 
\textbf{\textit{Average}} & \textbf{\textit{Median}}& \textbf{\textit{Range}}& \textbf{\textit{Mode}} \\
\hline
2.41 / 5                     & 2 / 5                      & 0 - 5                      & 2 / 5   \\
\hline
\end{tabular}
\label{survey_stats_mod}
\end{center}
\end{table}

Figure 2 shows the instances of a question being answered by a participant in the form of a 'confusion matrix'. The quadrants are determined based on whether the question being answered pertained to a modified explanation or an original one, and whether the participant agreed with the prediction or disagreed. The concentration of instances in the 'Unmodified-Agree' quadrant and their sparseness in the 'Unmodified-Disagree' quadrant indicates that explanations are effective in justifying predictions through highlighting most important features. The instances in the 'Modified-Disagree' quadrant are expected since random explanations conventionally should not justify the predictions. However, the high number of observations in 'Modified-Agree' shows that the participants often agree with the predictions due to wrong reasons, and are fooled by the explanation in this scenario. This implies that simple gradients in NLP are not effective for people assessing the accuracy of a model or understanding the internals of it.

\begin{figure}[h]%
    \centering
    \includegraphics[width=\linewidth]{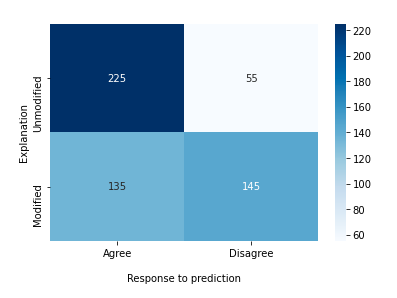}
    \caption{Confusion matrix of responses to explanations ()}
    \label{fig:my_label2}
\end{figure}

\section{Conclusions}
% What were the conclusions from the results
In this paper, we created a simple method to test the effectiveness of post-hoc explainability. The results of the survey indicate that simple gradient explanations in NLP are effective in helping end-users realize the most important features in predictions in a consumable form. However, the survey also shows that the participants often agree with the predictions due to wrong reasons, and are fooled by the explanation in this scenario. This implies that simple gradients in NLP are not effective for people in assessing the accuracy of a model or understanding the internals of it.
We can also conclude that current explainability methods that are based on gradients, specifically, simple gradients, might not clearly convey the reasons for the prediction to users. This will become more important as there is a need for trust and verifiability in models.

\section{Future Works}
%  recommend some future works
For future work, we propose that there should be more exploration of verifiable and trustworthy model explanations where humans can judge whether they are correct or incorrect. 
Proposed future works include:
\begin{itemize}
\item More work on verifiable and trustworthy model explanations.
\item A better way to provide decision support by using the natural language from the explanations.
\item Combining multiple models and providing explanations for the multiple models together.
\item Visualizing and using the explanations for debugging the models.
\item Providing explanations for the fairness of models.
\end{itemize}
    \bibliography{bibdata}
    \bibliographystyle{IEEEtran}

\end{document}